\documentclass[twoside,11pt]{article}

\usepackage{jmlr2e}
\usepackage{color}
\usepackage{subfigure}
\usepackage{epstopdf}
\usepackage{amsmath}
\usepackage{algorithmic}
\usepackage{algorithm}

\jmlrheading{EWRL}{2016}{0-0}{16/09}{03/12}{Mehdi Khamassi and Costas Tzafestas}

\ShortHeadings{Active exploration in parameterized reinforcement learning}{Khamassi and Tzafestas}
\firstpageno{1}

\begin{document}

\title{Active exploration in parameterized reinforcement learning}

\author{\name Mehdi Khamassi \email mehdi.khamassi@upmc.fr \\
       \addr Institute of Intelligent Systems and Robotics\\
       Sorbonne Universit\'es, UPMC Univ Paris 06, CNRS\\
       F-75005 Paris, France
       \AND
       \name Costas Tzafestas \email ktzaf@cs.ntua.gr \\
       \addr School of Electrical and Computer Engineering\\
       National Technical University of Athens\\
       Athens, Greece}

\maketitle

\begin{abstract}%   <- trailing '%' for backward compatibility of .sty file
Online model-free reinforcement learning (RL) methods with continuous actions are playing a prominent role when dealing with real-world applications such as Robotics. However, when confronted to non-stationary environments, these methods crucially rely on an exploration-exploitation trade-off which is rarely dynamically and automatically adjusted to changes in the environment. Here we propose an active exploration algorithm for RL in structured (parameterized) continuous action space. This framework deals with a set of discrete actions, each of which is parameterized with continuous variables. Discrete exploration is controlled through a Boltzmann softmax function with an inverse temperature $\beta$ parameter. In parallel, a Gaussian exploration is applied to the continuous action parameters. We apply a meta-learning algorithm based on the comparison between variations of short-term and long-term reward running averages to simultaneously tune $\beta$ and the width of the Gaussian distribution from which continuous action parameters are drawn. When applied to a simple virtual human-robot interaction task, we show that this algorithm outperforms continuous parameterized RL both without active exploration and with active exploration based on uncertainty variations measured by a Kalman-Q-learning algorithm.
\end{abstract}


\section{Introduction}

Important progresses have been made in recent years in reinforcement learning (RL) with continuous action spaces, permitting successful real-world applications such as Robotics applications \citep{kober11,stulp13}. Nevertheless, a recent review on reinforcement learning applied to Robotics \citep{kober13} highlighted, among other points, that (i) a variety of algorithms have been developed, each being appropriate to specific tasks: model-based versus model-free, function approximation versus policy search, continuous versus discrete action spaces; (ii) important human knowledge is injected concerning the search in the parameter space, either by reducing it through learning from demonstration, or by pre-adjusting parameters such as the exploration rate based on the prior determination of the total number of episodes in the experiment. In particular, the balance between exploration and exploitation is often pre-determined with human prior knowledge and does not extend well to tasks with non-stationary reward functions.

To address the first issue, the recent proposal of RL algorithms in structured Parameterized Action Space Markov Decision Processes (PAMDP) \citep{masson16,hausknecht16} seems to open a promising line of research. It combines a set of discrete actions $A_d = \{a_1,a_2,...,a_k\}$, each action $a \in A_d$ featuring $m_a$ continuous parameters $\{\theta_1^a,...,\theta_{m_a}^a\} \in \mathbb{R}^{m_a}$. Actions are thus represented by tuples $(a,\theta_1^a,...,\theta_{m_a}^a)$ and the overall action space is defined as $A = \cup_{a \in A_d} (a,\theta_1^a,...,\theta_{m_a}^a)$. This framework has been successfully applied to simulations of a Robocup 2D soccer task where agents have to learn to timely select between discrete actions such as running, turning or kicking the ball, and should learn at the same time with which speed to run, which angle to turn or which strength to kick. To ensure algorithm convergence, \citet{masson16} alternate between learning phases: (i) given a fixed policy for parameter selection, they use Q-Learning to optimize the policy discrete action selection; (ii) Next, they fix the policy for discrete action selection and use a policy search method to optimize the parameter selection. In contrast, \citet{hausknecht16} learn both in parallel by employing a parameterized actor that learns both discrete actions and parameters, and a parameterized critic that learns only the action-value function. Instead of relying on an external policy search procedure, they are thus able to directly query the critic for gradients.

Nevertheless, the exploration-exploitation trade-off is fixed in these methods, thus falling into the second issue raised by \citet{kober13}'s review. Exploration in continuous action spaces being different from discrete spaces, \citet{hausknecht16} adapt $\epsilon$-greedy exploration to parameterized action space by picking a random discrete action $a \in A_d$ with probability $\epsilon$ and sampling the action's parameters $\theta_i^a$ from a uniform random distribution. $\epsilon$ is arbitrarily annealed from 1.0 to 0.1 over the first 10,000 updates, thus requiring human prior knowledge about the duration of the task to appropriately tune exploration.

Here we use the Gaussian exploration for continuous action parameters proposed by \citet{vanHasselt07}, which in the original formulation uses a fixed Gaussian width $\sigma$. We then apply a noiseless version of the meta-learning algorithm of \citet{schweighofer03}, which tracks online variations of the agent's performance measured by short-term and long-term reward running averages. At each timestep, we use the difference between the two averages to simultaneously tune the inverse temperature $\beta_t$ used for selecting between discrete actions $a_j$, and the width $\sigma_t$ of the Gaussian distribution from which each continuous action parameter $\theta_i^a$ is sampled around its current value. We test our algorithm on a simple simulated human-robot interaction, where the algorithm tries to maximise reward computed as the virtual engagement of the human in the task, this engagement representing the attention that the human pays to the robot. We show that the proposed algorithm outperforms both continuous parameterized RL without active exploration and with active exploration based on uncertainty variations measured by a Kalman-RL algorithm \citep{geist10}.

\section{Active exploration algorithm}

The algorithm is summarized in Algorithm \ref{alg:algo}. It first employs Q-Learning \citep{watkins92} to learn the value of discrete action $a_t \in A_d$ selected at timestep $t$ in state $s_t$:

\begin{align}
Q_{t+1}(s_t,a_t) \leftarrow Q_{t}(s_t,a_t) + \alpha_Q [ r_t + \gamma \displaystyle\max_{a} (Q_t(s_{t+1},a)) - Q_t(s_t,a_t) ]
\label{eq:QL}
\end{align}

\noindent where $\alpha_Q$ is a learning rate and $\gamma$ is a discount factor. The probability of executing discrete action $a_j$ at timestep $t$ is given by a Boltzmann softmax equation:

\begin{align}
P(a_j | s_t, \beta_t) = \frac{exp\left( \beta_t Q_{t}(s_t,a_j) \right)}{\sum_a{exp\left( \beta_t Q_{t}(s_t,a) \right)}}
\label{eq:softmax}
\end{align}

\noindent where $\beta_t$ is a dynamic inverse temperature meta-parameter which will be tuned through meta-learning (see below).

In parallel, continuous $\widetilde{\theta}_{i,t}^{a_j}$ parameters with which action $a_j$ is executed at timestep $t$ are selected from a Gaussian exploration function centered on the current values $\theta_{i,t}^{a_j}(s_t)$ in state $s_t$ of the parameters of this action \citep{vanHasselt07}:

\begin{align}
P(\widetilde{\theta}_{i,t}^{a_j} | s_t, a_j, \sigma_t) = \frac{1}{\sqrt{2\pi\sigma_t}} exp \left( - (\widetilde{\theta}_{i,t}^{a_j} -  \theta_{i,t}^{a_j}(s_t))^2 / (2\sigma_t^2) \right)
\label{eq:gauss}
\end{align}

\noindent where the width $\sigma_t$ of the Gaussian is a meta-parameter which will be tuned through meta-learning (see below) and action parameters $\theta_{i,t}^a(s_t)$ are learned with a continuous actor-critic algorithm \citep{vanHasselt07}. A reward prediction error is computed from the critic: $\delta_t = r_t + \gamma V_t(s_{t+1}) - V_t(s_t)$ and is used to update the parameter vectors $\omega_{t}^C$ and $\omega_{t}^A$ of the neural network function approximations in the critic and the actor:

\begin{align}
\omega_{i,t+1}^C = \omega_{i,t}^C + \alpha_C \delta_t \frac{\delta V_t(s_{t})}{\delta \omega_{i,t}^C}
\text{ and }
\omega_{i,t+1}^A = \omega_{i,t}^A + \alpha_A \delta_t ( \widetilde{\theta}_{i,t}^a -  \theta_{i,t}^a(s_t)) \frac{\delta \theta_{i,t}^a(s_{t})}{\delta \omega_{i,t}^A}
\label{eq:critic}
\end{align}

\noindent where $\alpha_C$ and $\alpha_A$ are learning rates. In contrast to the original version where $\omega_{t}^A$ updates are performed only when $\delta_t > 0$ \citep{vanHasselt07}, here we update them all the time and proportionally to $\delta_t$ as in \citep{caluwaerts12}.

Finally, in order to perform active exploration on $\beta_t$ and $\sigma_t$, we implement a noiseless version of the meta-learning algorithm proposed by \citep{schweighofer03}. We compute short- and long-term reward running averages:

\begin{align}
\Delta \bar{r}(t) = (r(t) - \bar{r}(t)) / \tau_1
\text{ and } 
\Delta \bar{\bar{r}}(t) = (\bar{r}(t) - \bar{\bar{r}}(t)) / \tau_2
\label{eq:schweighofer}
\end{align}

\noindent where $\tau_1$ and $\tau_2$ are two time constants. We then update $\beta_t$ and $\sigma_t$ with:

\begin{align}
\beta_{t} = F(\mu (\bar{r}(t) - \bar{\bar{r}}(t)))
\text{ and } 
\sigma_{t} = G(\mu (\bar{r}(t) - \bar{\bar{r}}(t)))
\label{eq:meta}
\end{align}

\noindent where $\mu$ is a learning rate, $F(x)>0$ is affine, and $0<G(x)<20$ is a sigmoid.

We also compared this meta-learning algorithm with the Kalman Q-Learning proposed by \citep{geist10}. However, in contrast to the original formulation which proposes a purely exploratory agent by replacing Q-values in Equation \ref{eq:softmax} by the action-specific diagonal terms of the covariance matrix -- these terms representing the current variance/uncertainty about an action's Q-value --, here we multiply these terms by a weight $\eta$ and add them as exploration bonuses $b_t^a$ to Q-values in Equation \ref{eq:softmax}. We also use the covariance terms to tune action-specific $\sigma_t^a$ with function $G(x)$.

\begin{algorithm}
\caption{Active exploration with meta-learning\label{alg:algo}}
\begin{algorithmic}[1]
\STATE Initialize $\omega_{i,0}^A$, $\omega_{i,0}^C$, $Q_{i,0}$, $\beta_{0}$ and $\sigma_{0}$
\FOR {$t = 0,1,2,...$}
	\STATE Select discrete action $a_{t}$ with $softmax(s_{t},\beta_{t})$ (Eq. \ref{eq:softmax})
	\STATE Select action parameters $\widetilde{\theta}_{i,t}^a$ with $GaussianExploration(s_{t},a_{t},\theta_{i,t}^a,\sigma_{t})$ (Eq. \ref{eq:gauss})
	\STATE Observe new state and reward $\{s_{t+1},r_{t+1}\} \leftarrow Transition(s_t, a_t, \widetilde{\theta}_{i,t}^a)$
	\STATE Update  $Q_{t+1}(s_t,a_t)$ in the discrete Q-Learning (Eq. \ref{eq:QL})
	\STATE Update function approx. $\omega_{i,t+1}^C$ and $\omega_{i,t+1}^A$ in continuous actor-critic (Eq. \ref{eq:critic})
	\IF {meta-learning}
		\STATE Update reward running averages $\bar{r}(t)$ and $\bar{\bar{r}}(t)$ (Eq. \ref{eq:schweighofer})
		\STATE Update $\beta_{t}$ and $\sigma_{t}$ (Eq. \ref{eq:meta})
	\ENDIF
\ENDFOR
\end{algorithmic}
\end{algorithm}

\begin{figure} %[htbp]
  \begin{center}
%\floatconts
  {\includegraphics[width=1\textwidth]{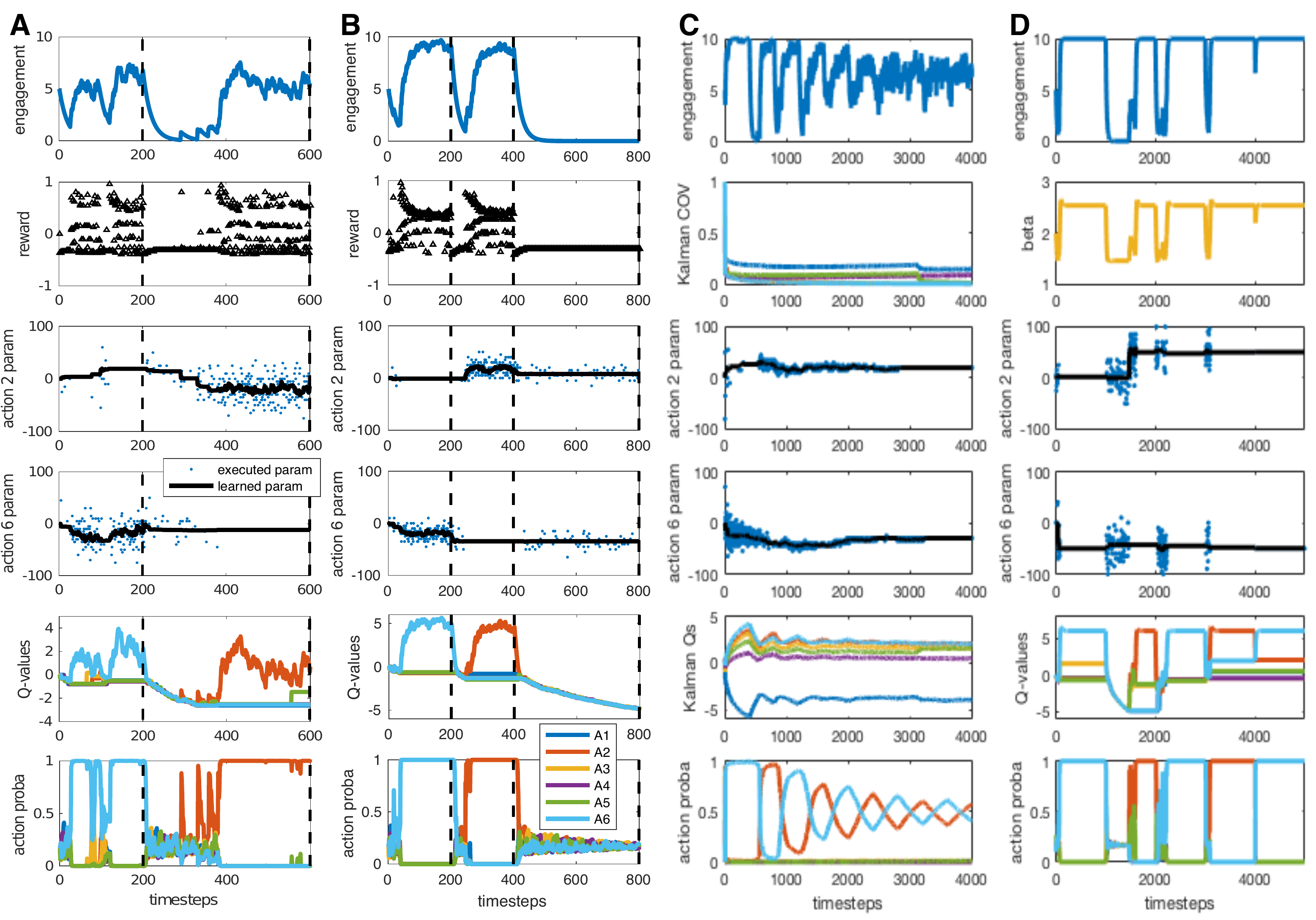}}
  {\caption{Simulations of the parameterized reinforcement learning (RL) algorithm with (A) fixed $\sigma=20$ and $\beta=4$ (no active exploration), (B) fixed $\sigma=10$ and $\beta=4$ (no active exploration), (C) $\sigma_t^a$ and $b_t^a$ tuned by Kalman-RL (active exploration) or (D) $\sigma_t$ and $\beta_t$ tuned by meta-learning (active exploration).}}% caption command
  \label{fig:fig1}% label
  \end{center}
\end{figure}

%Simulations of the reinforcement learning algorithm (A,B) without and (C,D) with active exploration.

\section{Experiments}

We test the algorithm described in Section 2 in a simple simulated human-robot interaction involving a single state, 6 discrete actions, and continuous action parameters between -100 and 100. The task is similar to a non-stationary stochastic multi-armed bandit task except that rather than associating a fixed probability of reward to each discrete action, an action will yield reward only when its continuous parameters are chosen within a Gaussian distribution around the current optimal action parameter $\mu^\star$ with variance $\sigma^\star$ (which are unknown to the robot). Every $n$ timesteps, $\mu^\star$ changes so that the task is non-stationary and requires constant re-exploration and learning by the robot.

The reward is given by a dynamical system which is based on the virtual engagement $e(t)$ of the human in the task. This engagement is supposed to represent the attention that the human pays to the robot. It starts at 5, increases up to a maximum $e_{max}=10$ when the robot performs the appropriate actions with the appropriate parameters, and decreases down to a minimum $e_{min}=0$ otherwise:

%\begin{align}
\[
e(t+1)= 
\begin{cases}
    e(t) + \eta_1 (e_{max} - e(t)) H(\theta_t^a),& \text{if } a(t)=a^\star \text{ \& } H(\theta_t^a)\geq0\\
    e(t) - \eta_2 (e_{min} - e(t)) H(\theta_t^a),& \text{if } a(t)=a^\star \text{ \& } H(\theta_t^a)<0\\
    e(t) + \eta_2 (e_{min} - e(t)),              & \text{otherwise}
\end{cases}
%\label{eq:eng}
\]
%\end{align}
where $\eta_1=0.1$ is the increasing rate, $\eta_2=0.05$ is the decreasing rate, and $H(x)$ is the reengagement function given by $H(x)=2\left(exp\left(- \frac{(x - \mu^\star)^2}{2\sigma^{\star2}}\right) - 0.5\right)$
where $a^\star$, $\mu^\star$ and $\sigma^\star$ are respectively the optimal action, action parameter and variance around $a^\star$.

The reward function is then computed as $r(t+1)=(1-\lambda)e(t+1)+\lambda \Delta e(t+1)$ where $\lambda=0.7$ is a weight. This reward function ensures that the algorithm gets rewarded in cases where the engagement $e(t+1)$ is low but nevertheless has just been increased by the action tuple $(a(t),\theta^a(t))$ performed by the robot.

%A) With sigma equal 20, the variance on action parameters is sufficiently large to adapt to environment changes, but engagement is suboptimal. B) With sigma equal 10, engagement is close to optimal when the task is learned, but the variance on action parameters is sometimes too small to adapt to environment changes. C) With sigma tuned by diagonal terms of the covariance matrix of the Kalman filter, the model progressively learns the average of two alternative conditions. D) With sigma and beta tuned by the difference between short- and long-term reward averages, the model performs transient increases in exploration following task changes and rapidly reconverge to optimal engagement.

We first simulated the algorithm without active exploration (thus with a fixed $\sigma=20$) in a task where the optimal action tuple ($a^\star$, $\mu^\star$) is $(a_6,-20)$ during 200 timesteps ($\sigma^\star=10$ in all the experiments presented here), then switches to $(a_2,-20)$ until timestep 600. Figure 1A shows that the algorithm first learns the appropriate action tuple $(a_6,-20)$, then takes some time to learn the second tuple, making the engagement drop between timesteps 200 and 400 and eventually finds the second optimal tuple. Nevertheless, $\sigma=20$ makes the robot select action parameters $\widetilde{\theta}_t^a$ with a large variance (illustrated by the clouds of blue dots around the learned action parameters $\theta_t^2$ and $\theta_t^6$ plotted as black curves). As a consequence, the engagement is not optimized and always remains below 7.5. In contrast, the same algorithm with a smaller fixed variance $\sigma=10$ can make the engagement reach the optimum of 10 when the optimal action tuple is learned (Figure 1B before timestep 400), but results in too little exploration which prevents the robot from finding a new action parameter which is too far away from the previously learned one (after timestep 400, the new optimal action tuple is $(a_6,20)$). These two examples illustrate the need to actively vary the variance $\sigma_t$ as a function of changes in the robot's performance.

We next tested active exploration with the Kalman Q-Learning algorithm in a task alternating between optimal tuples $(a_2,-20)$ and $(a_6,20)$ every 400 timesteps (Figure 1C). The diagonal terms of the covariance matrix $COV$ in the Kalman filter nearly monotonically decrease, resulting in a large variance $\sigma_t$ when action $a_6$ is executed until about timestep 600, and progressively decreasing the variance until the end of the experiment. Nevertheless, the algorithm quickly finds the appropriate action parameters and rapidly shifts between actions $a_2$ and $a_6$ after each change in the task condition. In the long run, the model progressively averages the statistics of the two conditions and learns to perform both actions with 50/50 probabilities (bottom part of Figure 1C) which decreases the simulated engagement (top).

We then tested active exploration with the meta-learning algorithm in a slightly more difficult task where the optimal action tuple alternate between $(a_2,-50)$ and $(a_6,50)$ every 1000 timesteps (Figure 1D). Transient drops in the engagement result in transient decreases in the exploration parameter $\beta_t$ as well as transient increases in the variance $\sigma_t$. This enables the algorithm to go through quick transient but wide exploration phases and to rapidly reconverge to exploitation, thus maximizing the simulated engagement.

Finally, we performed 10 simulations of each model on the difficult version of the task and plotted the average and standard deviation of the simulated engagement (Figure 2). The blue curve shows the performance of the algorithm without active exploration (i.e. fixed $\sigma=19$ obtained through parameter optimization), which adapts to each new condition but never exceeds a plateau of about 6. The green curve shows the active exploration with Kalman, which adapts faster at the beginning but progressively decreases its maximal engagement. The red curve shows the active exploration with meta-learning which initially takes more time to adapt but then only performs short transient explorations and reaches the optimum engagement of 10.

\begin{figure} %[htbp]
  \begin{centering}
%\floatconts
  {\includegraphics[width=0.7\textwidth]{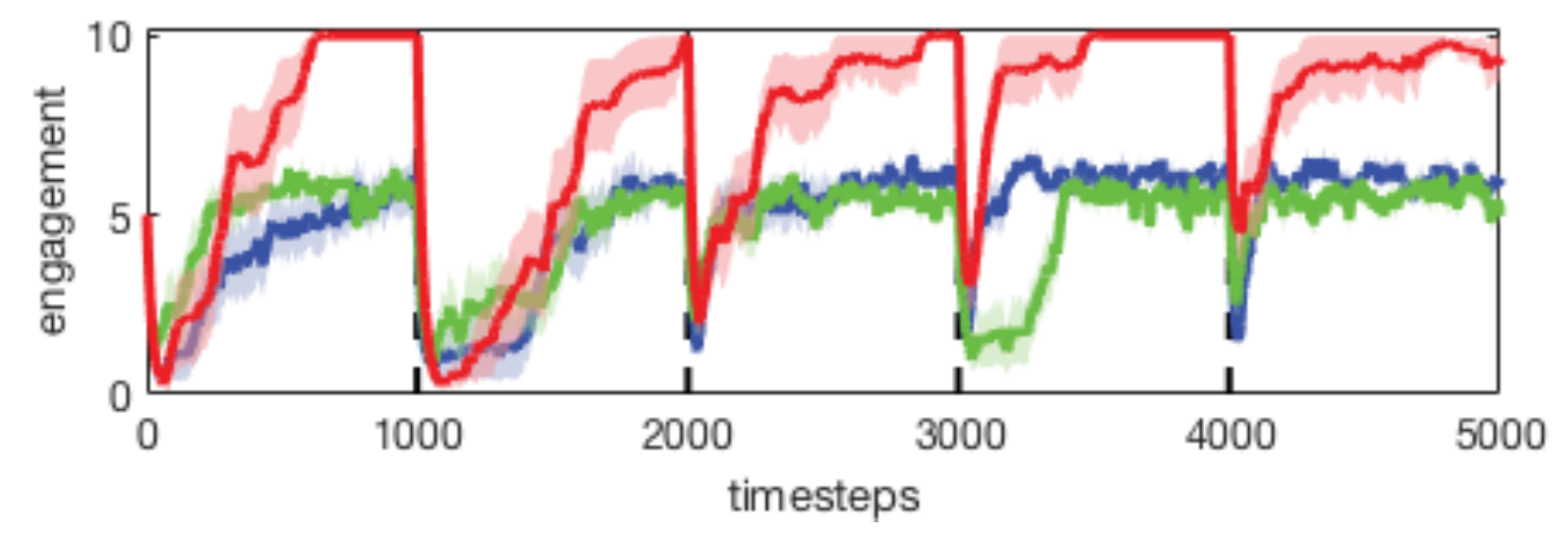}}
  {\caption{Comparison of engagement in 10 simulations of the meta-learning model (red), the model without active exploration (blue), and the Kalman-QL (green).}}% caption command
  \label{fig:fig2}% label
  \end{centering}
\end{figure}

\section{Conclusion}
  
In this work, we have shown that a meta-learning algorithm based on online variations of reward running averages can be used to adaptively tune two exploration parameters simultaneously used to select between both discrete actions and continuous action parameters in a parameterized action space. While we had previously successfully used the Kalman Q-Learning proposed by \citet{geist10} to coordinate model-based and model-free reinforcement learning in a stationary task \citep{viejo15}, it was not appropriate for the current non-stationary task. The proposed active exploration scheme could be a promising solution for Robotics applications of parameterized reinforcement learning.

\acks{We would like to thank Kenji Doya, Beno\^it Girard, Olivier Pietquin, Bilal Piot, Inaki Rano, Olivier Sigaud and Guillaume Viejo for useful discussions. This research work has been partially supported by the EU-funded Project BabyRobot (H2020-ICT-24-2015, grant agreement no. 687831) (MK, CT), by the Agence Nationale de la Recherche (ANR-11-IDEX-0004-02 Sorbonne-Universit\'es SU-15-R-PERSU-14 Robot Parallearning Project) (MK), and by Labex SMART (ANR-11- LABX-65 Online Budgeted Learning Project) (MK).}

\vskip 0.2in
\bibliography{biblio}

\begin{thebibliography}{11}
\providecommand{\natexlab}[1]{#1}
\providecommand{\url}[1]{\texttt{#1}}
\expandafter\ifx\csname urlstyle\endcsname\relax
  \providecommand{\doi}[1]{doi: #1}\else
  \providecommand{\doi}{doi: \begingroup \urlstyle{rm}\Url}\fi

\bibitem[Caluwaerts et~al.(2012)Caluwaerts, Staffa, S., Grand, Doll{\'e},
  Favre-F{\'e}lix, Girard, and Khamassi]{caluwaerts12}
K.~Caluwaerts, M.~Staffa, N'Guyen. S., C.~Grand, L.~Doll{\'e},
  A.~Favre-F{\'e}lix, B.~Girard, and M.~Khamassi.
\newblock A biologically inspired meta-control navigation system for the
  psikharpax rat robot.
\newblock \emph{Bioinspiration and Biomimetics}, 7\penalty0 (2):\penalty0
  025009, 2012.

\bibitem[Geist and Pietquin(2010)]{geist10}
M.~Geist and O.~Pietquin.
\newblock Kalman temporal differences.
\newblock \emph{Journal of artificial intelligence research}, 39:\penalty0
  483--532, 2010.

\bibitem[Hausknecht and Stone(2016)]{hausknecht16}
M.~Hausknecht and P.~Stone.
\newblock Deep reinforcement learning in parameterized action space.
\newblock In \emph{International Conference on Learning Representations (ICLR
  2016)}. 2016.

\bibitem[Kober and Peters(2011)]{kober11}
J.~Kober and J.~Peters.
\newblock Policy search for motor primitives in robotics.
\newblock \emph{Machine Learning}, 84:\penalty0 171--203, 2011.

\bibitem[Kober et~al.(2013)Kober, Bagnell, and Peters]{kober13}
J.~Kober, J.A. Bagnell, and J.~Peters.
\newblock Reinforcement learning in robotics: A survey.
\newblock \emph{The International Journal of Robotics Research}, pages
  1238--1274, 2013.
\newblock \doi{10.1177/0278364913495721}.

\bibitem[Masson and Konidaris(2016)]{masson16}
W.~Masson and G.~Konidaris.
\newblock Reinforcement learning with parameterized actions.
\newblock In \emph{Proceedings of the Thirtieth AAAI Conference on Artificial
  Intelligence (AAAI-16)}. 2016.

\bibitem[Schweighofer and Doya(2003)]{schweighofer03}
N.~Schweighofer and K.~Doya.
\newblock Meta-learning in reinforcement learning.
\newblock \emph{Neural Networks}, 16\penalty0 (1):\penalty0 5--9, 2003.

\bibitem[Stulp and Sigaud(2013)]{stulp13}
F.~Stulp and O.~Sigaud.
\newblock Robot skill learning: From reinforcement learning to evolution
  strategies.
\newblock \emph{Paladyn Journal of Behavioral Robotics}, 4\penalty0
  (1):\penalty0 49--61, 2013.

\bibitem[van Hasselt and Wiering(2007)]{vanHasselt07}
H.~van Hasselt and M.~Wiering.
\newblock Reinforcement learning in continuous action spaces.
\newblock In \emph{IEEE Symposium on Approximate Dynamic Programming and
  Reinforcement Learning}, pages 272--279. 2007.

\bibitem[Viejo et~al.(2015)Viejo, Khamassi, Brovelli, and Girard]{viejo15}
G.~Viejo, M.~Khamassi, A.~Brovelli, and B.~Girard.
\newblock Modeling choice and reaction time during arbitrary visuomotor
  learning through the coordination of adaptive working memory and
  reinforcement learning.
\newblock \emph{Frontiers in behavioral neuroscience}, 9, 2015.

\bibitem[Watkins and Dayan(1992)]{watkins92}
C.J.C.H. Watkins and P.~Dayan.
\newblock Q-learning.
\newblock \emph{Machine learning}, 8\penalty0 (3-4):\penalty0 279--292, 1992.

\end{thebibliography}

\end{document}